%% file: main.tex
\documentclass[11pt]{article}
\usepackage[preprint]{acl}
\usepackage{times}
\usepackage{latexsym}
\usepackage[T1]{fontenc}
\usepackage[utf8]{inputenc}
\usepackage{microtype}
\usepackage{inconsolata}
\hypersetup{
    colorlinks=true,
    linkcolor=blue,
    citecolor=blue,
    urlcolor=blue
}

\input{math_commands.tex}

\usepackage{url}
\usepackage{booktabs}
\usepackage{multirow}
\usepackage{placeins}
\usepackage{stfloats}
\usepackage{graphicx}
\usepackage{caption}
\usepackage{tabularx}
\usepackage{algorithm}
\usepackage{algpseudocode}
\newcommand{\algname}{ReAnchor}

\title{ReAnchor: Reusing Rejected Verifier States for Diffusion Drafting}

\author{
  \textbf{Sanghyun Lee}$^{1,*}$ \quad
  \textbf{Wonjin Nam}$^{1,*}$ \quad
  \textbf{Seungho Jang}$^{1}$ \quad
  \textbf{Jaeho Lee}$^{1}$ \\
  \textbf{Jiho Park}$^{1}$ \quad
  \textbf{Jaewoon Lee}$^{2}$ \quad
  \textbf{Seungryong Kim}$^{1,\dagger}$ \\[6pt]
  $^{1}$KAIST \qquad
  $^{2}$Seoul National University \\[4pt]
  \texttt{\{lsh83210,wjnam6525,jeffy1234,jihop00\}@kaist.ac.kr} \\
  \texttt{develop0235@gmail.com} \quad
  \texttt{theolee72@snu.ac.kr} \\
  \texttt{seungryong.kim@kaist.ac.kr}
}

\begin{document}

\maketitle
\begingroup
\renewcommand\thefootnote{}
\footnotetext{
$^{*}$ Equal contribution.
\quad
$^{\dagger}$ Corresponding author.
}
\endgroup
\begin{abstract}
Block diffusion drafters accelerate speculative decoding by generating multiple draft tokens in a single forward pass. 
However, each pass produces only a fixed-length block from a single target-generated anchor, limiting how far the draft can extend. 
Meanwhile, target verification computes hidden states even at positions beyond the first rejection, but these states are typically discarded rather than reused.
We propose \textbf{\algname{}}, a training-free method that reanchors a pretrained drafter at shifted positions using discarded verifier states to generate parallel draft blocks.
These blocks improve predictions within the original block and extend coverage beyond its endpoint.
At sampling temperature of $T=0$, \algname{} achieves its largest average gains on Gemma 4-12B, improving acceptance length by \textbf{19.4\%} and decoding speed by \textbf{17.9\%} across eight benchmarks. Averaged equally over four target models and eight benchmarks, it improves acceptance length by \textbf{15.2\%} and decoding speed by \textbf{13.4\%} with pretrained DFlash drafters. Our code is available at \href{https://github.com/cvlab-kaist/ReAnchor.git}{link}.
\end{abstract}

\section{Introduction}
Autoregressive large language models generate tokens sequentially, making decoding a major source of inference latency. Speculative decoding~\citep{leviathan2023fast, chen2023accelerating} reduces this cost by using a lightweight drafter to propose multiple tokens and verifying them in parallel with the target model.
Many existing methods use autoregressive drafters~\citep{li2026eagle, cheng2024recurrent, li2024eagle2}, whose token-by-token generation introduces sequential work into the drafting stage. Block diffusion drafters such as DFlash~\citep{chen2026dflash} remove this sequential dependence by predicting multiple draft tokens in a single forward pass.

This single-pass block generation comes at the cost of flexibility in how draft proposals are used and extended. Unlike autoregressive drafters, which can extend proposals one token at a time and stop when further drafting is unlikely to be worthwhile~\citep{huang2024specdec, zhang2025draft}, block diffusion drafters generate a fixed-length block determined by their training configuration. When verification encounters an early rejection, hidden states computed for the remaining draft positions are typically discarded. Conversely, when the entire block is accepted, no further proposals are available beyond its endpoint. Simply increasing the block length at inference time forces the drafter outside its training configuration and degrades acceptance performance~\citep{chen2026dflash}. Thus, block diffusion drafting leaves verifier computations underutilized when proposals are rejected early, while limiting draft coverage when the entire block is accepted.

Our observation is that verifier hidden states computed at discarded draft positions retain useful information for subsequent drafting (Table~\ref{tab:post_rejection_agreement}). Although these states depend on rejected tokens and cannot be reused as valid target-model states, they can still serve as conditioning signals for the pretrained drafter. This suggests that otherwise discarded verifier computations can be leveraged for additional drafting without further target-model evaluations.

\begin{figure}[t]
  \includegraphics[width=\linewidth]{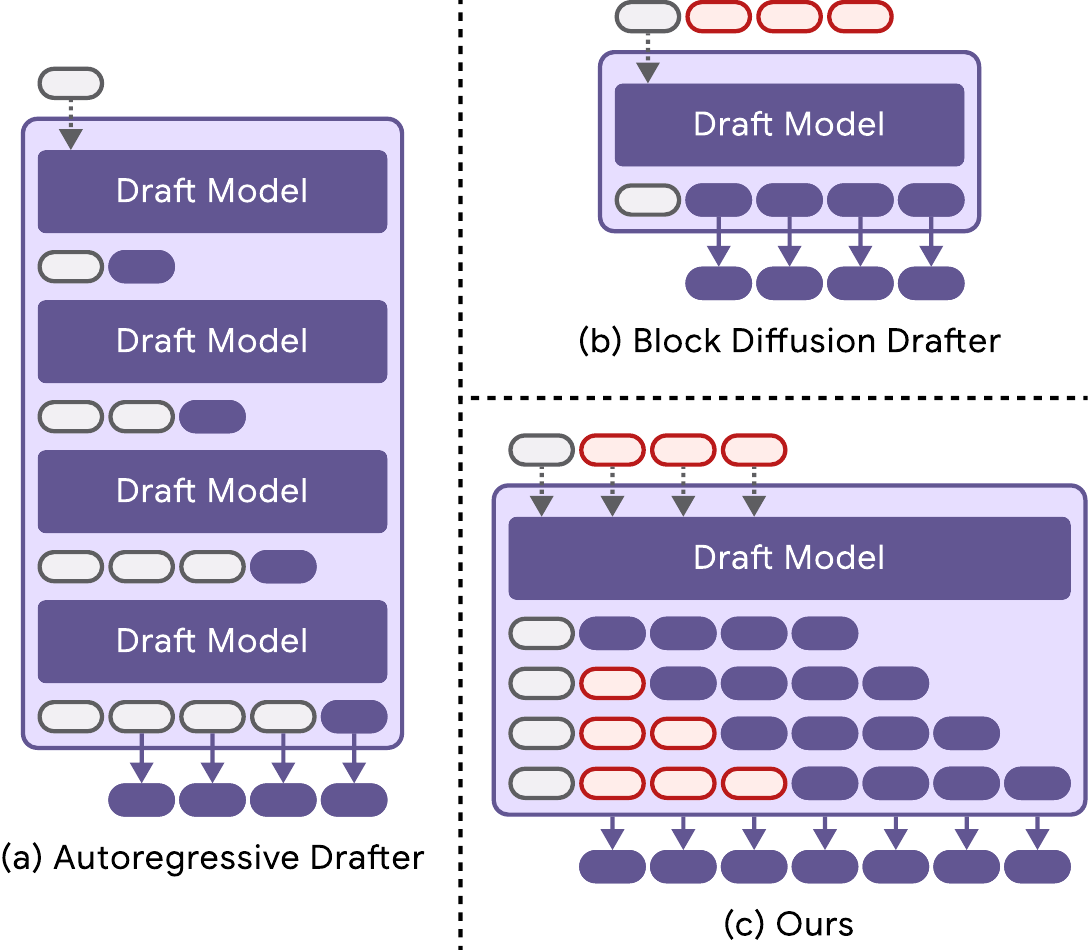}
  \caption{
    \textbf{Comparison of Drafting Strategies.} While (a) autoregressive drafting requires multiple sequential next-token predictions and (b) block diffusion drafting generates a single draft block in one forward pass, (c) ReAnchor generates multiple draft blocks through parallel, low-cost drafter forward passes, reusing verifier hidden states discarded after rejection in the previous round to draft from later positions.
  }
  \label{fig:placeholder}
\end{figure}

Building on this finding, we introduce \textbf{\algname{}}, a training-free method that reuses verifier hidden states from the discarded post-rejection suffix to reanchor the same pretrained drafter at those later positions. The reanchored blocks are generated in parallel with the baseline draft, each preserving the drafter's trained block length. To combine these blocks, \algname{} matches their starting tokens to the baseline and attaches their continuations at the matched positions. By reanchoring at later positions (Figure~\ref{fig:placeholder}), these blocks provide alternative predictions for nearby tokens within the original block, while also extending draft coverage beyond its endpoint. All attached continuations are verified by the target model under their new prefixes, preserving lossless decoding.

With DFlash drafters, \algname{} improves average acceptance length by \textbf{15.2\%} and decoding speed by \textbf{13.4\%} across four target models and eight benchmarks at $T=0$. The largest average gains are observed on Gemma 4-12B, with improvements of \textbf{19.4\%} in acceptance length and \textbf{17.9\%} in decoding speed. These improvements also extend to three DFlash-based drafter variants, Domino, DSpark, and Draft-OPD, which employ different training and prediction mechanisms. Across 48 comparisons spanning two Qwen3 target models and eight benchmarks, \algname{} improves average acceptance length by \textbf{11.8\%} and end-to-end decoding speed by \textbf{9.8\%} over their respective baselines. Component-wise ablations further show that both in-block refinement and block extension contribute to these gains.

Our contributions are summarized as follows:
\begin{itemize}
  \item We introduce \algname{}, a training-free method that reuses
    verifier hidden states as shifted anchors for block diffusion drafting.

  \item \algname{} improves DFlash by refining predictions within the original block and extending draft coverage beyond it, achieving consistent gains in acceptance length and decoding speed across diverse settings.

    \item We demonstrate that \algname{} can be directly applied to three DFlash-based variants, Domino, DSpark, and Draft-OPD, without additional training or variant-specific algorithmic modifications.
\end{itemize}

\section{Preliminaries}
We review DFlash-style block diffusion drafting and speculative decoding, introducing the notation and key concepts used throughout this work. Related work is discussed in Appendix~\ref{related_work}.
\label{sec:prelim}
\subsection{Block Diffusion Drafter}

A block diffusion drafter generates multiple draft tokens in a single forward pass.  The drafter receives two inputs: an \emph{anchor token}, i.e., the
bonus token that the autoregressive target model produces at the last
verified position as a byproduct of its verification pass, and
\emph{prefix context} derived from the target model's hidden states
over the accepted prefix. Given these inputs, the drafter simultaneously predicts a fixed number of tokens at the positions immediately following the anchor.

We now formalize this procedure.
Let \(p\) denote the target model and \(q\) the diffusion
drafter.  We write \(B\) for the \emph{block size}: one anchor
followed by \(B-1\) drafted tokens.  All positions are absolute;
\(i\) denotes the anchor position. The anchor token \(a_i=x_i\sim p(\cdot\mid x_{<i})\) is the
target-generated bonus token, where \(x_{<i}\) is the accepted
prefix. The target model's hidden states
\(\mathbf{H}_{<i}=(\mathbf{h}_j)_{j<i}\) are projected into the
drafter's hidden space by a learned linear projection
\(\Pi_\phi\) to form the prefix context
\(\mathbf{C}_{<i}=\Pi_\phi(\mathbf{H}_{<i})\).
The drafter takes the anchor followed by \(B-1\) mask tokens
\(\mathbf{m}_{i+1:i+B-1}\) and produces a distribution over each
masked position:
\[
  \mathbf{z}^{[i]}=(a_i,\;\mathbf{m}_{i+1:i+B-1}),\quad
  \mathbf{q}^{[i]}_j=q(\mathbf{C}_{<i};\;\mathbf{z}^{[i]})_j,
\]
\vspace{-1.2em}
\[
  \hat{x}^{[i]}_j\sim\operatorname{Cat}(\mathbf{q}^{[i]}_j),
  \quad j=i{+}1,\ldots,i{+}B{-}1.
\]
The superscript \([i]\) identifies the anchor.

\subsection{Speculative Decoding}

Given a draft block $\hat{\mathbf{x}}^{[i]}_{i+1:i+B-1}$, the target
model $p$ evaluates all proposed positions in a single forward pass.
Verification then proceeds from left to right, accepting draft tokens
as long as they match tokens independently sampled from the target
distribution at each position. At the first mismatch, verification
stops and accepts the target-sampled token.

Unlike standard probabilistic verification, which uses importance
ratios and residual sampling, this sample-wise verification does
not require evaluating proposal probabilities. It preserves the
target distribution because each accepted token is sampled from
$p$ under its verified prefix.

Let $\tau$ denote the average number of tokens accepted
per round, including the target-generated token at the stopping
position. The average decoding latency per accepted token is
\[
    L = \frac{T_{\mathrm{draft}} + T_{\mathrm{verify}}}{\tau}.
\]
The drafter therefore aims to generate low-cost proposals that
match the target's sampled trajectory for as many consecutive
positions as possible.
\begin{figure}[t]
  \includegraphics[width=\linewidth]{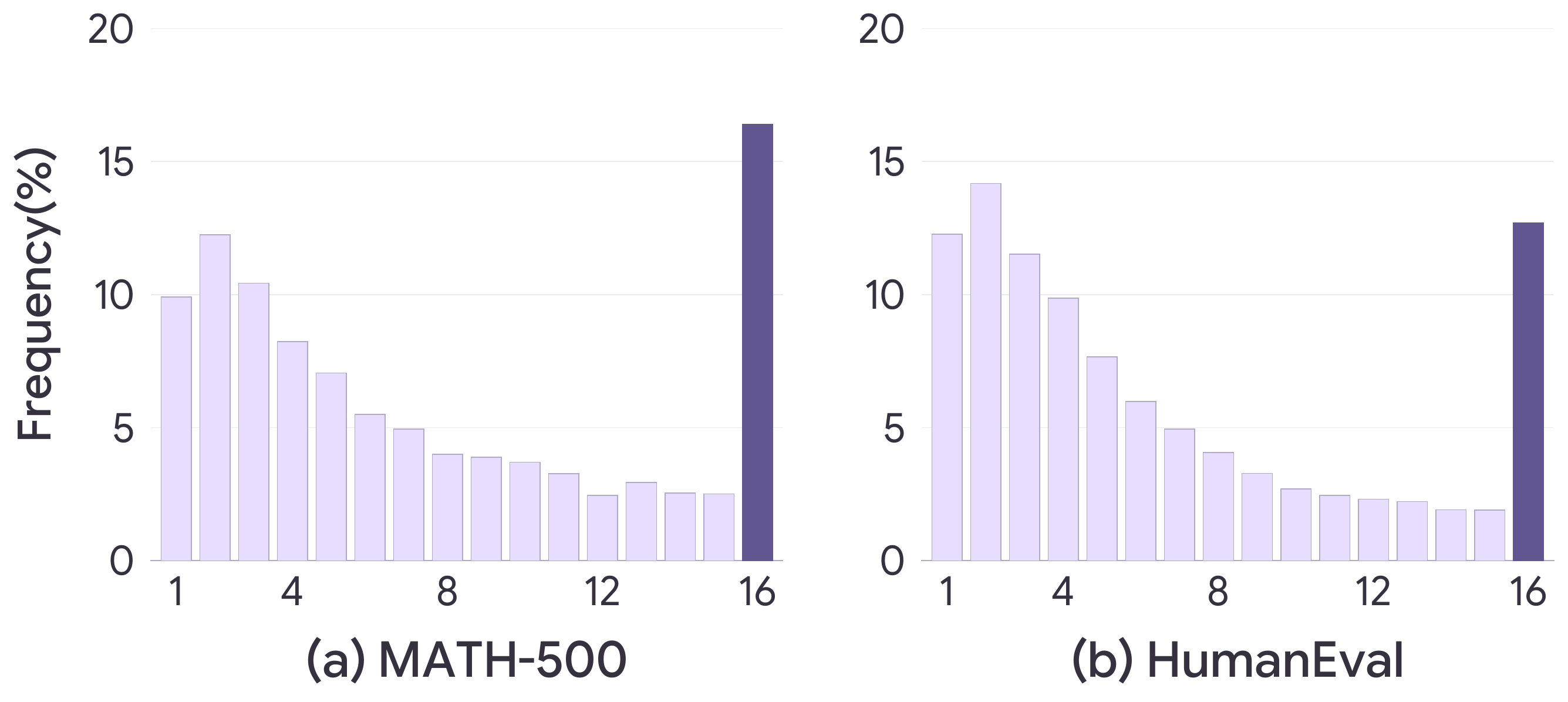}
  \caption{\textbf{Distribution of per-round acceptance length on (a) MATH-500 and (b) HumanEval.} The spike at position 16 indicates frequent saturation at the draft-block boundary.}
  \label{fig:applength}
\end{figure}
\section{Method}
\subsection{Revisiting Diffusion Drafting}
\label{sec:motivation}
Block diffusion drafting removes the sequential bottleneck of autoregressive drafters, but the fixed-length parallel block introduces constraints that do not arise in the sequential setting. We now revisit this formulation and examine where its fixed-block design leaves room for improvement.  We present two observations that together motivate \algname{}:
(i)~the native block endpoint is frequently the binding constraint
on acceptance length and no proposal exists beyond it, yet naively
extending the block degrades acceptance;
(ii)~verifier states computed over rejected draft suffixes retain
substantial predictive information and can condition additional
drafts without retraining.

\paragraph{Block-End Saturation and Naive Extension.}
\label{par:saturation}
The native draft length frequently limits the accepted prefix, but
simply increasing the proposal length at inference time does not
resolve this issue. Figure~\ref{fig:applength} shows the acceptance-length
distribution on representative math and code tasks for drafters
trained with \(B=16\).  Both distributions exhibit a clear
concentration at the maximum length. All drafted tokens are accepted in \textbf{16.4\%} of
rounds on MATH-500 and \textbf{12.7\%} on HumanEval. In these rounds, every drafted position is consumed and no further proposal
is available. Full-block acceptance does not establish that the next token would also be accepted, but the concentration at the final bin shows that the block endpoint is the binding constraint in a substantial fraction of decoding rounds.

A direct solution is to increase the proposal length at inference time. We test this by evaluating drafters trained with \(B=16\) at \(B=24\) and \(B=32\) on Qwen3-8B across MATH-500, GSM8K, HumanEval, and MBPP. Despite the additional positions, the average accepted length decreases by \textbf{0.3} and \textbf{1.1} respectively compared to the native block size. The longer block forces the drafter to predict at offsets beyond its training range from the same anchor, under a masked input configuration it was not trained on. Extending coverage therefore motivates while preserving the native block size.  We next examine whether the rejected portion of the preceding verification pass provides the conditioning information needed at these later starting points.

\begin{table}[tbp]
  \centering
  \small
  \setlength{\tabcolsep}{3pt}
  \renewcommand{\arraystretch}{1.06}
  \begin{tabular*}{\linewidth}{
      @{\extracolsep{\fill}}@{}lcccc@{}
    }
    \toprule
    & GSM8K & HumanEval & MBPP & MATH-500 \\
    \midrule
    Agreement (\%)
    & 65.4 & 60.2 & 60.6 & 66.0 \\
    \bottomrule
  \end{tabular*}
  \caption{\textbf{Post-rejection agreement (\%).} Fraction of draft
    tokens after the first rejection that match the target model's
    top-1 predictions.}
  \label{tab:post_rejection_agreement}
\end{table}

\paragraph{Predictive Information after Rejection.}
\label{par:rejected_info}
A rejection at position \(r\) prevents the remaining draft suffix
from being accepted, but does not render every subsequent draft
token incorrect.  Table~\ref{tab:post_rejection_agreement} reports
the fraction of draft tokens after \(r\) that match the target
model's top-1 prediction.  Across the evaluated tasks,
\textbf{60--66\%} of these agree with despite
following a rejected prefix.  This indicates that the suffix is
invalidated as a contiguous sequence rather than on a per-token
basis, and that the discarded positions still carry substantial
predictive information.  Since the target model has already computed
hidden states at these positions during verification, these states
are available as conditioning signals for subsequent drafting at no additional target-model cost. Motivated by these observations, the next subsection introduces \textbf{\algname{}}, which reuses these rejected-prefix verifier states as shifted anchors for the same pretrained drafter to generate additional fixed-length proposals.

\begin{figure*}[t]
  \centering
  \includegraphics[width=\textwidth]{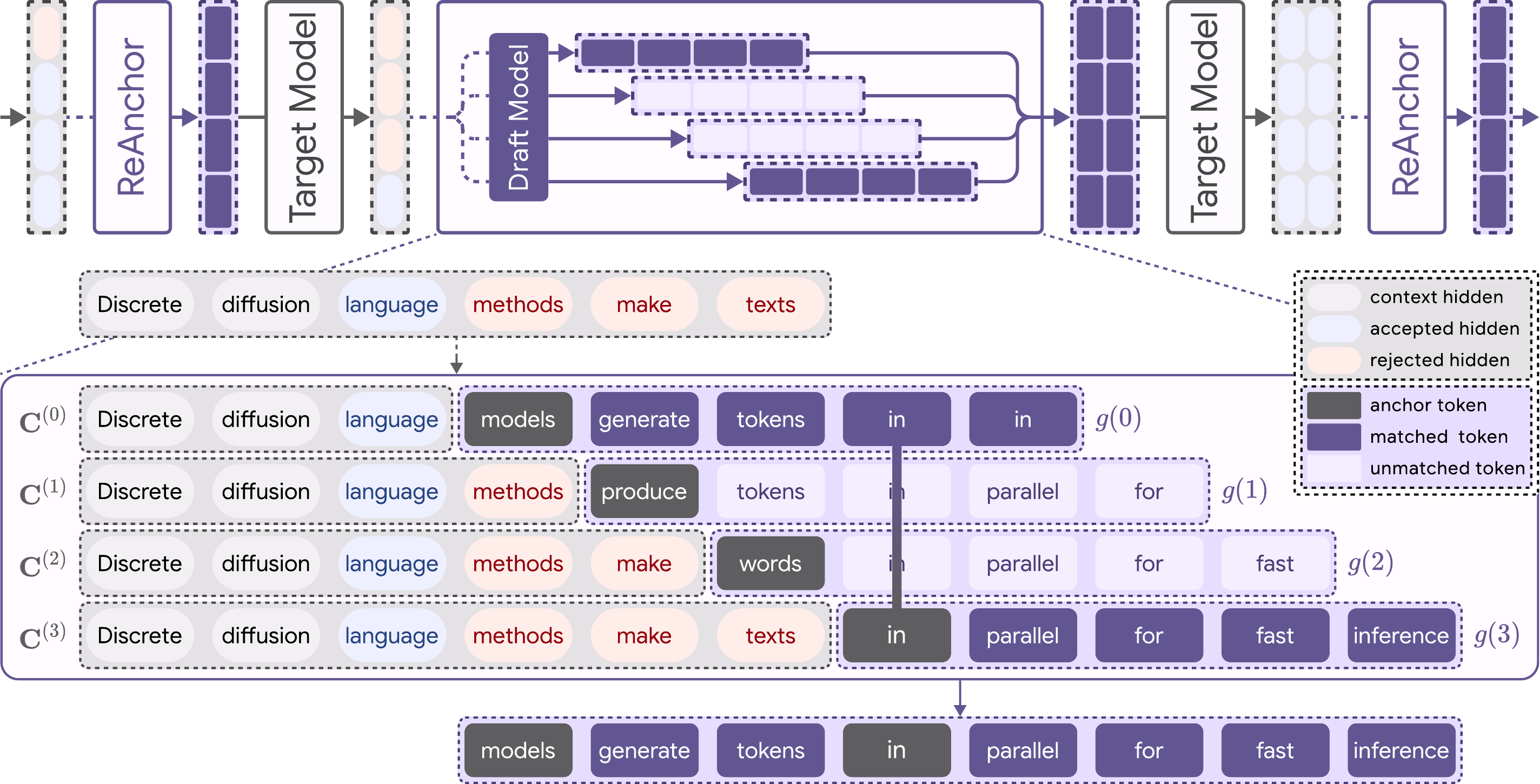}
  \caption{
    \textbf{Overview of \algname{}'s Draft Construction.}
    (top) ReAnchor is a training-free speculative decoding method
    using pretrained block-diffusion drafter \(q\). At each
    decoding round, ReAnchor generates draft and target model
    \(p\) verifies it repeatedly. (bottom) Rejected draft tokens of
    previous round from target model are typically discarded at
    prior methods. However, ReAnchor leverages them by directly
    conditioning them into pretrained draft model \(q\).
    Specifically, ReAnchor generates
    \((B-1)(1+|\mathcal D|)\) draft tokens with parallel inference of
    draft model \(q\). Each group of \(B-1\) draft tokens is
    generated from a different anchor
    position. We can construct the final proposal structure 
    by matching and attaching generated draft tokens.
  }
  \label{fig:placeholder_method}
\end{figure*}

\subsection{\algname{}}
\label{sec:method}

\algname{} constructs draft proposals by combining the standard baseline
draft with additional shifted blocks conditioned on verifier hidden
states discarded after rejection in the preceding round
(Figure~\ref{fig:placeholder_method}). These drafts are generated in
parallel using the same pretrained drafter at its trained block length.
Each shifted block is aligned with the baseline by matching its starting
token, and its continuation is attached at the matched position. The
resulting paths are merged into a token trie and jointly verified by the
target model. This construction provides alternative predictions within
the original block while extending draft coverage beyond its endpoint.

\paragraph{Parallel drafting from discarded verifier states.}
Section~\ref{sec:motivation} shows that verifier states from rejected
prefixes remain useful for drafting. \algname{} reuses a subset of these
states as conditioning points for the same pretrained drafter.
The resulting shifted blocks improve predictions within the original
block and extend the draft beyond its endpoint without retraining the
drafter.

The baseline drafter uses the anchor \(a_i\) and \(\mathbf C_{<i}\) to generate \(\hat{\mathbf x}^{[i]}_{i+1:i+B-1}\). Let \(i'\) and \(r\) denote the anchor and the first rejected position of the preceding round; the token accepted at \(r\) becomes the current anchor, so \(i=r\).
We retain the otherwise discarded verifier states \(\widetilde{\mathbf h}_j\) over the rejected suffix \(r\le j\le i'+B-1\), and let \(\widetilde{x}_j\) denote the token decoded from \(\widetilde{\mathbf h}_{j-1}\) through the target LM head.
Let
\[
\mathcal D\subseteq\{\,d\in\{1,\ldots,B-1\}: i+d\le i'+B\,\}
\]
denote the selected offsets from the current anchor.
For each \(d\in\mathcal D\), we define the shifted anchor \(\widetilde a_{i+d}=\widetilde x_{i+d}\) and its context \(\widetilde{\mathbf C}_{<i+d}=\Pi_\phi(\widetilde{\mathbf H}_{<i+d})\), where \(\widetilde{\mathbf H}_{<i+d}=(\mathbf H_{<r}, \widetilde{\mathbf h}_r,\ldots,\widetilde{\mathbf h}_{i+d-1})\), mirroring the baseline pair \((a_i,\mathbf C_{<i})\).
The same pretrained drafter takes \(\widetilde a_{i+d}\) followed by \(B-1\) masked positions and predicts them in parallel. The baseline and shifted blocks are
\begin{align*}
g(0) &= \bigl(a_i,\hat{\mathbf x}^{[i]}_{i+1:i+B-1}\bigr),\\
g(d) &= \bigl(\widetilde a_{i+d},
        \hat{\mathbf x}^{[i+d]}_{i+d+1:i+d+B-1}\bigr),
\end{align*}
each consisting of one anchor and \(B-1\) proposed tokens.
The entry \(g(d)_k\), \(k\in\{0,\ldots,B-1\}\), nominally corresponds
to absolute position \(i+d+k\).

The \(1+|\mathcal D|\) drafting calls are independent and can be batched together. Each block attends only to its own verifier context and masked input, and reusing states from the preceding verification pass requires no additional target evaluation. 
The subset \(\mathcal D\) controls the number of additional drafting calls; in this work, we use every available offset, \(\mathcal D=\{1,\ldots,i'+B-i\}\). When the preceding round accepts its entire block, no states are retained, \(\mathcal D\) is empty, and only the baseline draft is generated.

\paragraph{Token matching and continuation attachment.}
\algname{} produces one baseline block and several shifted blocks, which must be combined into a single proposal set. A shifted block \(g(d)\) is drafted from a reused anchor at offset \(d\), so it would naturally be appended to the baseline at that position. However, correcting the rejected token changes the continuation that follows it, so the tokens drafted in \(g(d)\) may now appear at a different position in the new baseline \(g(0)\), or not at all. Rather than fixing \(g(d)\) at its original offset \(d\), we therefore search the baseline for a position at which \(g(d)\) aligns and attach its continuation there.

We say that \(g(d)\) has a positive overlap at baseline position \(t\) if its anchor matches the baseline token there, i.e., \(g(0)_t=g(d)_0\). We attach \(g(d)\) at the earliest such position \(t^\star\), so that its continuation follows the baseline prefix \(g(0)_{0:t^\star}\) (Algorithm~\ref{alg:reanchor}). This gives every matched shifted block a unique, deterministic attachment point without a secondary ranking rule, and in practice it selects a path with high accepted length (Appendix~\ref{app:earliest_attachment}).

The overlapping prefix is represented only once in the concatenated path, and shifted blocks with no positive-overlap match are discarded; each shifted block therefore contributes at most one attached path. We collect the baseline and at most one attachment per shifted block into \(\mathcal Y\) and merge them into a token trie.  Since a token match only identifies a candidate attachment rather than validating its continuation, we jointly verify the trie in a single target forward pass using tree attention~\citep{miao2024specinfer}, so that each continuation is evaluated under its attached prefix.
\begin{algorithm}[tbp]
  \small
  \caption{\algname{} Proposal Construction}
  \label{alg:reanchor}
  \algrenewcommand{\algorithmicindent}{1em}
  \begin{algorithmic}[1]
    \Require Target context \(\mathbf C_{<i}\), anchor \(a_i\),
    block size \(B\)
    \Require Retained verifier states \(\widetilde{\mathbf H}\),
    source tokens \(\widetilde{\mathbf x}\),
    selected offsets \(\mathcal D\) (Sec.~\ref{sec:method})
    \Ensure Proposal set \(\mathcal Y\) for target verification

    \State \(\mathbf C^{(0)}\gets\mathbf C_{<i},
    \quad g(0)_0\gets a_i\)
    \ForAll{\(d\in\mathcal D\)}
    \State \(
      \begin{aligned}[t]
        \mathbf C^{(d)}&\gets\Pi_\phi(\widetilde{\mathbf H}_{<i+d}),\\
        g(d)_0&\gets\widetilde{a}_{i+d}
    \end{aligned}\)
    \EndFor

    \ForAll{\(d\in\{0\}\cup\mathcal D\) \textbf{in parallel}}
    \State \(\mathbf z^{[i+d]}
    \gets(g(d)_0,\mathbf m_{i+d+1:i+d+B-1})\)
    \State \(
      \begin{aligned}[t]
        &(\mathbf q_j^{[i+d]})_{j=i+d+1}^{i+d+B-1}\\
        &\quad\gets q(\mathbf C^{(d)};\mathbf z^{[i+d]})
    \end{aligned}\)
    \State Generate \(g(d)_{1:B-1}\) from these distributions
    using the chosen drafting rule.
    \EndFor

    \State \(\mathcal Y\gets\{g(0)\}\)
    \ForAll{\(d\in\mathcal D\)}
    \State \(\mathcal T\gets\{t\in\{0,\ldots,B-1\}: g(0)_t=g(d)_0\}\)
    \If{\(\mathcal T\neq\{\}\)}
      \State \(t^\star\gets\min\mathcal T\)
      \State \(\mathcal Y\gets\mathcal Y\cup
      \{(g(0)_{0:t^\star-1},g(d)_{0:B-1})\}\)
    \EndIf
    \EndFor
    \State Merge identical token prefixes in \(\mathcal Y\) into a trie.
    \State \Return \(\mathcal Y\)
  \end{algorithmic}
\end{algorithm}

\section{Experiments}
\subsection{Experimental setup}
\paragraph{Models and Evaluation.}
We evaluate \algname{} with Qwen3~\citep{yang2025qwen3} models at
4B and 8B scales, Llama-3.1-8B~\citep{grattafiori2024llama} and Gemma 4-12B~\citep{gemmateam2026gemma4technicalreport}. Experiments cover
GSM8K~\citep{cobbe2021training}, MATH-500 ~\citep{lightman2024let}, AIME25, HumanEval~\citep{chen2021evaluating}, MBPP
~\citep{austin2021program}, LiveCodeBench
~\citep{jain2025livecodebench}, MT-Bench ~\citep{zheng2023judging},
and Alpaca ~\citep{taori2023alpaca}. We report average acceptance
length \(\tau\) and end-to-end decoding speedup relative to
autoregressive decoding with the target model. We evaluate on 100 examples per dataset, sampled with random seed~0, using the full set for AIME25 (30) and MT-Bench (80). In all experiments, \algname{} uses every retained rejected-suffix state as a shifted anchor.

\begin{table*}[tp]
  \centering
  \resizebox{\textwidth}{!}{%
  \begin{tabular}{@{}clllllllll@{}}
    \toprule
    \multicolumn{2}{c}{} & \multicolumn{8}{c}{\textbf{Dataset}} \\
    \cmidrule(lr){3-10}
    \textbf{Target model} & \textbf{Drafting method}
      & GSM8K & MATH & AIME25 & HumanEval & MBPP
      & \begin{tabular}[b]{@{}c@{}}LiveCode\\Bench\end{tabular}
      & MT-Bench & Alpaca \\
    \midrule
    \multicolumn{10}{l}{\textbf{Temperature = 0}} \\
    \midrule
    \multirow{2}{*}{Qwen3-4B} & DFlash
      & 6.24 / 4.19$\times$ & 7.37 / 5.10$\times$ & 6.36 / 4.35$\times$ & 6.51 / 4.39$\times$ & 5.81 / 4.13$\times$ & 5.87 / 4.19$\times$ & 3.46 / 2.44$\times$ & 2.89 / 2.01$\times$ \\
    & + \algname{}
      & \textbf{7.51} / \textbf{5.03$\times$} & \textbf{8.45} / \textbf{5.78$\times$} & \textbf{7.20} / \textbf{4.90$\times$} & \textbf{7.50} / \textbf{5.00$\times$} & \textbf{6.78} / \textbf{4.78$\times$} & \textbf{6.84} / \textbf{4.76$\times$} & \textbf{3.89} / \textbf{2.68$\times$} & \textbf{3.17} / \textbf{2.17$\times$} \\
    \midrule
    \multirow{2}{*}{Qwen3-8B} & DFlash
      & 6.34 / 4.39$\times$ & 7.28 / 5.05$\times$ & 6.38 / 4.41$\times$ & 6.49 / 4.43$\times$ & 5.71 / 4.06$\times$ & 5.94 / 4.10$\times$ & 3.42 / 2.40$\times$ & 2.87 / 1.99$\times$ \\
    & + \algname{}
      & \textbf{7.45} / \textbf{5.03$\times$} & \textbf{8.31} / \textbf{5.67$\times$} & \textbf{7.11} / \textbf{4.72$\times$} & \textbf{7.52} / \textbf{5.05$\times$} & \textbf{6.68} / \textbf{4.69$\times$} & \textbf{7.24} / \textbf{4.83$\times$} & \textbf{3.82} / \textbf{2.57$\times$} & \textbf{3.16} / \textbf{2.14$\times$} \\
    \midrule
    \multirow{2}{*}{Gemma 4-12B} & DFlash
      & 3.88 / 3.27$\times$ & 4.97 / 4.14$\times$ & 4.56 / 3.80$\times$ & 4.48 / 3.70$\times$ & 4.12 / 3.37$\times$ & 4.07 / 3.27$\times$ & 2.96 / 2.37$\times$ & 2.77 / 2.16$\times$ \\
    & + \algname{}
      & \textbf{4.79} / \textbf{3.99$\times$} & \textbf{6.04} / \textbf{4.94$\times$} & \textbf{5.30} / \textbf{4.39$\times$} & \textbf{5.75} / \textbf{4.69$\times$} & \textbf{5.04} / \textbf{4.05$\times$} & \textbf{4.74} / \textbf{3.69$\times$} & \textbf{3.36} / \textbf{2.65$\times$} & \textbf{3.13} / \textbf{2.48$\times$} \\
    \midrule
    \multirow{2}{*}{Llama-3.1-8B} & DFlash
      & 4.42 / 2.71$\times$ & 3.85 / 2.43$\times$ & 3.82 / 2.43$\times$ & 5.03 / 3.26$\times$ & 5.02 / 3.13$\times$ & 4.02 / 2.51$\times$ & 3.85 / 2.42$\times$ & 3.70 / 2.32$\times$ \\
    & + \algname{}
      & \textbf{4.98} / \textbf{3.11$\times$} & \textbf{4.25} / \textbf{2.65$\times$} & \textbf{4.25} / \textbf{2.63$\times$} & \textbf{5.60} / \textbf{3.58$\times$} & \textbf{5.65} / \textbf{3.52$\times$} & \textbf{4.44} / \textbf{2.69$\times$} & \textbf{4.28} / \textbf{2.62$\times$} & \textbf{4.17} / \textbf{2.59$\times$} \\
    \midrule
    \multicolumn{10}{l}{\textbf{Temperature = 1}} \\
    \midrule
    \multirow{2}{*}{Qwen3-4B} & DFlash
      & 4.90 / 3.54$\times$ & 4.81 / 3.32$\times$ & 3.70 / 2.62$\times$ & 5.00 / 3.60$\times$ & 4.56 / 3.26$\times$ & 4.68 / \textbf{3.34$\times$} & 2.66 / 1.86$\times$ & 2.30 / 1.63$\times$ \\
    & + \algname{}
      & \textbf{5.64} / \textbf{3.80$\times$} & \textbf{5.53} / \textbf{3.73$\times$} & \textbf{4.24} / \textbf{2.83$\times$} & \textbf{5.69} / \textbf{3.90$\times$} & \textbf{5.16} / \textbf{3.49$\times$} & \textbf{4.93} / 3.18$\times$ & \textbf{2.86} / \textbf{1.94$\times$} & \textbf{2.46} / \textbf{1.67$\times$} \\
    \midrule
    \multirow{2}{*}{Qwen3-8B} & DFlash
      & 4.62 / 3.36$\times$ & 4.68 / 3.32$\times$ & 3.57 / 2.53$\times$ & 4.42 / 3.20$\times$ & 4.03 / 2.93$\times$ & 4.72 / 3.37$\times$ & 2.43 / 1.75$\times$ & 2.23 / 1.58$\times$ \\
    & + \algname{}
      & \textbf{5.35} / \textbf{3.79$\times$} & \textbf{5.11} / \textbf{3.46$\times$} & \textbf{3.94} / \textbf{2.71$\times$} & \textbf{4.88} / \textbf{3.34$\times$} & \textbf{4.52} / \textbf{3.20$\times$} & \textbf{5.42} / \textbf{3.73$\times$} & \textbf{2.63} / \textbf{1.77$\times$} & \textbf{2.38} / \textbf{1.63$\times$} \\
    \midrule
    \multirow{2}{*}{Gemma 4-12B} & DFlash
      & 4.39 / 3.76$\times$ & 4.88 / 4.12$\times$ & 4.17 / 3.45$\times$ & 4.19 / 3.46$\times$ & 3.79 / 3.12$\times$ & 3.62 / 3.00$\times$ & 2.86 / 2.22$\times$ & 2.59 / 1.81$\times$ \\
    & + \algname{}
      & \textbf{5.45} / \textbf{4.48$\times$} & \textbf{5.84} / \textbf{4.71$\times$} & \textbf{4.92} / \textbf{3.92$\times$} & \textbf{5.24} / \textbf{4.18$\times$} & \textbf{4.70} / \textbf{3.72$\times$} & \textbf{4.22} / \textbf{3.24$\times$} & \textbf{3.22} / \textbf{2.38$\times$} & \textbf{2.94} / \textbf{2.04$\times$} \\
    \midrule
    \multirow{2}{*}{Llama-3.1-8B} & DFlash
      & 2.92 / \textbf{1.80$\times$} & 2.31 / \textbf{1.52$\times$} & \textbf{1.93} / \textbf{1.32$\times$} & 3.58 / 2.19$\times$ & 3.53 / 2.31$\times$ & 2.68 / 1.68$\times$ & 2.38 / \textbf{1.47$\times$} & 2.43 / 1.49$\times$ \\
    & + \algname{}
      & \textbf{2.97} / 1.76$\times$ & \textbf{2.40} / 1.49$\times$ & \textbf{1.93} / 1.18$\times$ & \textbf{3.87} / \textbf{2.34$\times$} & \textbf{3.86} / \textbf{2.49$\times$} & \textbf{2.95} / \textbf{1.73$\times$} & \textbf{2.48} / 1.46$\times$ & \textbf{2.63} / \textbf{1.57$\times$} \\
    \bottomrule
  \end{tabular}}
  \caption{\textbf{DFlash decoding performance at sampling temperatures
    $T=0$ and $T=1$.} Each entry reports average acceptance length $\tau$ /
    end-to-end speedup. Temperature $T$ is used for both drafter and target
    sampling in speculative decoding. Speedup is the ratio of target-only
    latency measured at the same temperature to speculative decoding latency;
    $\tau$ includes the target bonus token produced at each verification round.
    Bold marks the best value per dataset and metric.}
  \label{tab:dflash_acceptance_length}
\end{table*}

\paragraph{Drafter Setup.}
Unless otherwise specified, our main experiments and ablations use pretrained DFlash~\citep{chen2026dflash} checkpoints. We additionally apply \algname{} to three DFlash-based variants to assess its compatibility with different training and prediction mechanisms. Draft-OPD~\citep{lei2026draftopdonpolicydistillationspeculative} post-trains DFlash through on-policy distillation with error-position replay. Domino~\citep{huang2026domino} adds a GRU-based causal correction head. DSpark~\citep{cheng2026dspark} uses a lightweight Markov head for sequential token prediction. Further details on the drafter variants are provided in Appendix~\ref{variant_dflash}. 

\paragraph{Latency Measurement.}
We measure end-to-end wall-clock decoding latency with batch size one on a single NVIDIA H100 GPU. All compared methods use the same target model, drafter checkpoint, hardware, and generation settings. We exclude model loading and tokenization time while including measured latencies for both drafting and target verification for end-to-end latency evaluation. GPU execution is synchronized before timing, and we report results after warm-up runs. More details of the setup are provided in Appendix~\ref{exp_detail}.

\subsection{Main Results}
\label{sec:main_results}

\paragraph{Acceptance length and decoding speed.}
At $T=0$, \algname{} improves both average acceptance length $\tau$ and end-to-end speedup in all 32 target--benchmark comparisons, achieving average gains of \textbf{15.2\%} and \textbf{13.4\%}, respectively (Table~\ref{tab:dflash_acceptance_length}). The largest relative gains occur on Gemma 4-12B, with improvements of \textbf{19.4\%} in $\tau$ and \textbf{17.9\%} in speedup, despite its shorter baseline accepted prefixes than Qwen3. At $T=1$, the improvements remain substantial, with $\tau$ increasing in 31 of 32 comparisons and speedup in 27, averaging \textbf{11.8\%} and \textbf{6.9\%}, respectively. The smaller gains under stochastic sampling may stem from increased token mismatches between independently sampled draft and target continuations, reducing the benefit of additional proposals relative to their verification cost. Overall, these results demonstrate the effectiveness of reanchoring across different target models and decoding settings.

\paragraph{Compatibility with drafter variants.}
We further evaluate \algname{} on three DFlash-based variants, Domino, DSpark, and Draft-OPD, with different training and prediction mechanisms. At $T=0$, \algname{} improves both $\tau$ and end-to-end speedup in all 48 comparisons, with average gains of \textbf{11.8\%} and \textbf{9.8\%}, respectively (Table~\ref{tab:other_drafter_acceptance_length}). At $T=1$, the improvements remain consistent, increasing $\tau$ in all 48 comparisons and speedup in 45, with average gains of \textbf{11.5\%} and \textbf{8.1\%}. These results demonstrate that \algname{} generalizes across different drafting mechanisms without additional training or variant-specific modifications.

\begin{table*}[tp]
  \centering
  \renewcommand{\arraystretch}{1.1}
  \resizebox{\textwidth}{!}{%
  \begin{tabular}{@{}clllllllll@{}}
    \toprule
    \multicolumn{2}{c}{} & \multicolumn{8}{c}{\textbf{Dataset}} \\
    \cmidrule(lr){3-10}
    \textbf{Target model} & \textbf{Drafting method}
      & GSM8K & MATH & AIME25 & HumanEval & MBPP
      & \begin{tabular}[b]{@{}c@{}}LiveCode\\Bench\end{tabular}
      & MT-Bench & Alpaca \\
    \midrule
    \multicolumn{10}{l}{\textbf{Temperature = 0}} \\
    \midrule
    \multirow{6}{*}[-4pt]{Qwen3-4B} & Domino
      & 9.32 / 5.82$\times$ & 8.20 / 5.34$\times$ & 6.49 / 4.19$\times$ & 6.84 / 4.50$\times$ & 6.57 / 4.42$\times$ & 5.82 / 3.83$\times$ & 4.06 / 2.71$\times$ & 3.56 / 2.33$\times$ \\
      & + \algname{}
      & \textbf{11.01} / \textbf{6.84$\times$} & \textbf{9.31} / \textbf{5.94$\times$} & \textbf{7.46} / \textbf{4.72$\times$} & \textbf{7.73} / \textbf{4.99$\times$} & \textbf{7.75} / \textbf{5.09$\times$} & \textbf{6.75} / \textbf{4.29$\times$} & \textbf{4.48} / \textbf{2.93$\times$} & \textbf{3.90} / \textbf{2.50$\times$} \\
    \cmidrule(lr){2-10}
      & DSpark
      & 6.42 / 4.38$\times$ & 6.20 / 4.27$\times$ & 5.52 / 3.85$\times$ & 5.70 / 4.10$\times$ & 5.49 / 4.07$\times$ & 5.18 / 3.67$\times$ & 3.98 / 2.90$\times$ & 3.61 / 2.46$\times$ \\
      & + \algname{}
      & \textbf{6.81} / \textbf{4.72$\times$} & \textbf{6.47} / \textbf{4.41$\times$} & \textbf{5.87} / \textbf{4.05$\times$} & \textbf{6.11} / \textbf{4.35$\times$} & \textbf{5.93} / \textbf{4.36$\times$} & \textbf{5.62} / \textbf{4.00$\times$} & \textbf{4.30} / \textbf{3.11$\times$} & \textbf{3.96} / \textbf{2.65$\times$} \\
    \cmidrule(lr){2-10}
      & Draft-OPD
      & 7.10 / 4.82$\times$ & 8.27 / 5.76$\times$ & 6.76 / 4.75$\times$ & 6.66 / 4.73$\times$ & 6.23 / 4.48$\times$ & 6.10 / 4.34$\times$ & 3.54 / 2.57$\times$ & 2.94 / 2.08$\times$ \\
      & + \algname{}
      & \textbf{8.35} / \textbf{5.66$\times$} & \textbf{9.46} / \textbf{6.50$\times$} & \textbf{7.78} / \textbf{5.33$\times$} & \textbf{7.64} / \textbf{5.33$\times$} & \textbf{7.27} / \textbf{5.17$\times$} & \textbf{7.09} / \textbf{4.90$\times$} & \textbf{3.91} / \textbf{2.78$\times$} & \textbf{3.20} / \textbf{2.22$\times$} \\
    \midrule
    \multirow{6}{*}[-4pt]{Qwen3-8B} & Domino
      & 9.60 / 6.32$\times$ & 8.39 / 5.38$\times$ & 6.60 / 4.20$\times$ & 7.20 / 4.72$\times$ & 6.69 / 4.37$\times$ & 6.22 / 4.09$\times$ & 4.06 / 2.56$\times$ & 3.56 / 2.28$\times$ \\
      & + \algname{}
      & \textbf{11.01} / \textbf{7.08$\times$} & \textbf{9.50} / \textbf{6.00$\times$} & \textbf{7.46} / \textbf{4.62$\times$} & \textbf{8.32} / \textbf{5.29$\times$} & \textbf{7.87} / \textbf{5.06$\times$} & \textbf{6.95} / \textbf{4.22$\times$} & \textbf{4.60} / \textbf{2.80$\times$} & \textbf{4.02} / \textbf{2.50$\times$} \\
    \cmidrule(lr){2-10}
      & DSpark
      & 6.54 / 4.73$\times$ & 6.30 / 4.45$\times$ & 5.52 / 3.75$\times$ & 5.94 / 4.14$\times$ & 5.56 / 3.94$\times$ & 5.47 / 3.68$\times$ & 3.96 / 2.71$\times$ & 3.71 / 2.54$\times$ \\
      & + \algname{}
      & \textbf{6.95} / \textbf{5.07$\times$} & \textbf{6.59} / \textbf{4.60$\times$} & \textbf{5.94} / \textbf{3.97$\times$} & \textbf{6.29} / \textbf{4.42$\times$} & \textbf{5.98} / \textbf{4.18$\times$} & \textbf{5.87} / \textbf{3.97$\times$} & \textbf{4.29} / \textbf{2.89$\times$} & \textbf{4.03} / \textbf{2.73$\times$} \\
    \cmidrule(lr){2-10}
      & Draft-OPD
      & 7.23 / 5.24$\times$ & 8.31 / 5.70$\times$ & 6.93 / 4.81$\times$ & 6.67 / 4.71$\times$ & 6.23 / 4.45$\times$ & 6.11 / 4.24$\times$ & 3.52 / 2.45$\times$ & 2.92 / 1.94$\times$ \\
      & + \algname{}
      & \textbf{8.46} / \textbf{5.99$\times$} & \textbf{9.33} / \textbf{6.27$\times$} & \textbf{7.88} / \textbf{5.33$\times$} & \textbf{7.71} / \textbf{5.32$\times$} & \textbf{7.18} / \textbf{5.01$\times$} & \textbf{7.19} / \textbf{4.77$\times$} & \textbf{3.87} / \textbf{2.61$\times$} & \textbf{3.21} / \textbf{2.13$\times$} \\
    \midrule
    \multicolumn{10}{l}{\textbf{Temperature = 1}} \\
    \midrule
    \multirow{6}{*}[-4pt]{Qwen3-4B} & Domino
      & 7.46 / 5.53$\times$ & 5.66 / 4.22$\times$ & 4.18 / 3.15$\times$
      & 5.59 / 4.17$\times$ & 5.21 / 3.89$\times$ & 4.77 / 3.14$\times$
      & 3.06 / 2.36$\times$ & 2.72 / 2.07$\times$ \\
      & + \algname{}
      & \textbf{8.63} / \textbf{6.16$\times$} & \textbf{6.47} / \textbf{4.68$\times$}
      & \textbf{4.79} / \textbf{3.49$\times$} & \textbf{6.44} / \textbf{4.67$\times$}
      & \textbf{6.04} / \textbf{4.38$\times$} & \textbf{5.52} / \textbf{3.50$\times$}
      & \textbf{3.32} / \textbf{2.46$\times$} & \textbf{3.01} / \textbf{2.21$\times$} \\
      \cmidrule(lr){2-10}
      & DSpark
      & 5.20 / 4.15$\times$ & 4.58 / 3.61$\times$ & 3.76 / 3.04$\times$
      & 4.53 / 3.62$\times$ & 4.42 / 3.30$\times$ & 4.80 / 3.43$\times$
      & 3.09 / 2.58$\times$ & 3.04 / 2.49$\times$ \\
      & + \algname{}
      & \textbf{5.78} / \textbf{4.55$\times$} & \textbf{4.99} / \textbf{3.99$\times$}
      & \textbf{3.99} / \textbf{3.15$\times$} & \textbf{5.02} / \textbf{3.89$\times$}
      & \textbf{4.91} / \textbf{3.67$\times$} & \textbf{5.25} / \textbf{3.92$\times$}
      & \textbf{3.36} / \textbf{2.73$\times$} & \textbf{3.23} / \textbf{2.54$\times$} \\
      \cmidrule(lr){2-10}
      & Draft-OPD
      & 5.49 / 4.63$\times$ & 5.40 / 4.58$\times$ & 3.95 / 3.38$\times$
      & 5.16 / 4.32$\times$ & 4.83 / 4.10$\times$ & 4.79 / 3.57$\times$
      & 2.76 / 2.39$\times$ & 2.39 / 2.04$\times$ \\
      & + \algname{}
      & \textbf{6.48} / \textbf{5.25$\times$} & \textbf{6.04} / \textbf{4.93$\times$}
      & \textbf{4.64} / \textbf{3.81$\times$} & \textbf{6.00} / \textbf{4.82$\times$}
      & \textbf{5.58} / \textbf{4.55$\times$} & \textbf{5.41} / \textbf{3.86$\times$}
      & \textbf{2.94} / \textbf{2.44$\times$} & \textbf{2.56} / \textbf{2.10$\times$} \\
      \midrule
      \multirow{6}{*}[-4pt]{Qwen3-8B} & Domino
      & 7.13 / 4.56$\times$ & 5.88 / 3.71$\times$ & 4.20 / 2.69$\times$
      & 5.38 / 3.43$\times$ & 4.83 / 3.09$\times$ & 4.93 / 3.05$\times$
      & 2.98 / 1.89$\times$ & 2.72 / 1.71$\times$ \\
      & + \algname{}
      & \textbf{8.09} / \textbf{5.02$\times$} & \textbf{6.66} / \textbf{4.04$\times$}
      & \textbf{4.81} / \textbf{2.94$\times$} & \textbf{5.98} / \textbf{3.67$\times$}
      & \textbf{5.70} / \textbf{3.52$\times$} & \textbf{5.82} / \textbf{3.41$\times$}
      & \textbf{3.27} / \textbf{1.98$\times$} & \textbf{3.01} / \textbf{1.80$\times$} \\
      \cmidrule(lr){2-10}
      & DSpark
      & 5.42 / 4.07$\times$ & 4.78 / 3.45$\times$ & 4.00 / \textbf{2.93$\times$}
      & 4.92 / 3.68$\times$ & 4.74 / 3.65$\times$ & 4.60 / \textbf{3.21$\times$}
      & 3.32 / 2.37$\times$ & 3.09 / 2.14$\times$ \\
      & + \algname{}
      & \textbf{5.97} / \textbf{4.43$\times$} & \textbf{5.14} / \textbf{3.59$\times$}
      & \textbf{4.05} / 2.92$\times$ & \textbf{5.31} / \textbf{3.99$\times$}
      & \textbf{5.27} / \textbf{3.95$\times$} & \textbf{4.85} / 3.16$\times$
      & \textbf{3.64} / \textbf{2.60$\times$} & \textbf{3.34} / \textbf{2.33$\times$} \\
      \cmidrule(lr){2-10}
      & Draft-OPD
      & 5.17 / 3.74$\times$ & 5.06 / 3.55$\times$ & 3.86 / 2.82$\times$
      & 4.71 / 3.42$\times$ & 4.39 / 3.16$\times$ & 5.02 / 3.55$\times$
      & 2.52 / 1.81$\times$ & 2.27 / \textbf{1.60$\times$} \\
      & + \algname{}
      & \textbf{6.24} / \textbf{4.35$\times$} & \textbf{5.76} / \textbf{3.94$\times$}
      & \textbf{4.24} / \textbf{2.95$\times$} & \textbf{5.17} / \textbf{3.61$\times$}
      & \textbf{4.96} / \textbf{3.43$\times$} & \textbf{5.77} / \textbf{3.83$\times$}
      & \textbf{2.73} / \textbf{1.86$\times$} & \textbf{2.37} / 1.60$\times$ \\
    \bottomrule
  \end{tabular}}
  \caption{\textbf{Decoding performance of DFlash-based drafter variants at $T=0$ and $T=1$.}
    Each entry reports average acceptance length $\tau$ / end-to-end speedup relative
    to target-only decoding at the same temperature. The accepted length includes
    the target bonus token. Bold marks the best value per dataset and metric.}
  \label{tab:other_drafter_acceptance_length}
\end{table*}

\begin{table}[!htbp]
  \centering
  \small
  \renewcommand{\arraystretch}{1.06}

  \begin{tabular*}{\columnwidth}{
    @{\extracolsep{\fill}}@{}lccc@{}
  }
    \toprule
    Method & In-block & Beyond & $\Delta\tau$ \\
    \midrule
    DFlash
    & $\times$ & $\times$ & -- \\

    In-block only
    & $\checkmark$ & $\times$ & $+0.54$ \\

    Extension only
    & $\times$ & $\checkmark$ & $+0.31$ \\
    \bottomrule
  \end{tabular*}

  \caption{\textbf{In-block refinement vs.\ block extension.}
    $\Delta\tau$ over DFlash. \textit{In-block only} uses shifted
    predictions within the native block; \textit{Extension only}
    uses them only beyond its endpoint.}
  \label{tab:ablation_region}
\end{table}

\subsection{Ablation Study}
We analyze where the gains of \algname{} come from: which draft positions benefit, how much is due to reusing rejected states versus token matching, and how the attachment rule affects performance. All ablations use Qwen3-4B with the DFlash drafter at \(T=0\). \paragraph{In-Block Refinement and Block Extension.} We separate the gains of \algname{} into improvements within the native draft block and extensions beyond its endpoint. We consider two partial variants while keeping the same shifted anchors and drafter evaluations. \textit{In-block only} uses predictions from shifted anchors only at positions up to \(i+B-1\) and discards all predictions beyond the native block. \textit{Extension only} retains the original DFlash predictions within \(i+1:i+B-1\) and uses \algname{} only to provide additional predictions after \(i+B-1\). The full method uses both.
\begin{table}[tbp]

  \centering
  \small

  \setlength{\tabcolsep}{3pt}
  \renewcommand{\arraystretch}{1.06}

  \begin{tabular*}{\linewidth}{
      @{\extracolsep{\fill}}@{}lcc@{}
    }
    \toprule
    Task & Shifted only & Multiple \\
    \midrule
    GSM8K     & 50.7 & 35.8 \\
    MATH-500   & 51.2 & 40.6 \\
    HumanEval & 49.0 & 34.7 \\
    MBPP      & 45.5 & 38.0 \\
    \bottomrule
  \end{tabular*}
  \caption{
  \textbf{Start-token alignment (\%).}
  \emph{Shifted only}: the anchor matches only at baseline positions
  other than its original offset. \emph{Multiple}: it matches at more
  than one position. Categories may overlap.
}
  \label{tab:token_alignment}
\end{table}

\begin{table}[t]
\centering
\small
\resizebox{\columnwidth}{!}{%
\begin{tabular}{lcc}
\toprule
Attachment rule & Acceptance $\tau$ & Speedup \\
\midrule
      Original offset      & 7.24 & $4.96\times$ \\
      Random match         & 7.51 & $5.16\times$ \\
      Earliest match       & \textbf{7.56} & \textbf{5.16}$\times$ \\
      Longest-prefix match & 7.51 & $5.14\times$ \\
\bottomrule
\end{tabular}%
}
\caption{\textbf{Ablation of attachment-site matching on Qwen3-4B with DFlash.} Results are averaged over 100 examples each from GSM8K, HumanEval, MATH, and MBPP.}
\label{tab:attachment_matching}
\end{table}
\begin{table*}[!tbp]
  \centering
  \small
  \setlength{\tabcolsep}{5pt}
  \renewcommand{\arraystretch}{1.0}

  \begin{tabular*}{\textwidth}{@{\extracolsep{\fill}}@{}lcccccc@{}}
    \toprule
    Configuration & MATH & GSM8K & MBPP & HumanEval & MT-Bench& Alpaca  \\
    \midrule
    DFlash
    &7.37 & 6.24 & 5.81 & 6.51 & 3.46 & 2.89 \\
    \quad + Fixed-offset ReAnchor
    & 8.03 & 6.89 & 6.49 & 7.04 & 3.75 & 3.04 \\
    \quad + \algname{}
    & \textbf{8.45} & \textbf{7.51} & \textbf{6.78}
    & \textbf{7.50} & \textbf{3.89} & \textbf{3.17} \\
    \bottomrule
  \end{tabular*}
    \caption{\textbf{Cumulative ablation of \algname{} components.}
    Average acceptance length \(\tau\) on Qwen3-4B with the DFlash
    drafter at \(T=0\). \textit{Fixed-offset} attaches each shifted
    block at its original offset \(d\); \algname{} instead attaches it at
    the earliest matching position in the baseline draft. All
    configurations use the same shifted anchors and drafter evaluations.}
  \label{tab:ablation_reuse_matching}
\end{table*}
Table~\ref{tab:ablation_region} shows that both components improve over the DFlash baseline. In-block refinement increases acceptance by providing later positions with predictions conditioned from closer verifier states. Block extension provides additional accepted tokens in rounds where verification reaches the native endpoint. Combining the two gives the largest improvement in both acceptance length and decoding speed, showing that \algname{} benefits from improving existing draft positions as well as extending the draft beyond its original length.

\paragraph{Analysis of token alignment.}
We next examine how often a shifted block \(g(d)\) remains aligned with its original offset \(d\) after a rejection. Correcting the rejected token can alter the continuation that follows it, so the anchor \(g(d)_0\) may appear at a different position of the new baseline draft \(g(0)\). Table~\ref{tab:token_alignment} reports two statistics over shifted blocks whose anchor matches at least one position \(t\), i.e., \(g(0)_t=g(d)_0\). \textit{Shifted only} counts blocks that do not match at \(t=d\) but match at some \(t\neq d\); \textit{Multiple} counts blocks that match at more than one \(t\). The two categories may overlap.

Across the evaluated tasks, \textbf{45.5--51.2\%} of matched shifted blocks fall into the shifted-only category, indicating that enforcing the original offset would discard a substantial fraction of otherwise usable continuations. In addition, \textbf{34.7--40.6\%} admit multiple candidate attachment positions, showing that alignment is frequently ambiguous rather than fixed. These results motivate searching over baseline positions when attaching shifted blocks. We next quantify the effect of this design choice by comparing fixed-offset attachment with the full token-matching strategy of \algname{}.

\paragraph{Ablation of attachment-site matching.}
We compare our default earliest-overlap rule with fixed-offset, random-match, and longest-prefix attachment on Qwen3-4B with DFlash (Table~\ref{tab:attachment_matching}). All matching rules clearly outperform fixed-offset attachment (7.51--7.56 vs.\ 7.24 in \(\tau\)), while differing from one another by at most 0.05 in \(\tau\) and 0.02\(\times\) in speedup. Re-alignment itself thus matters more than the specific attachment site. We adopt the earliest rule because it matches the best results while being the simplest: a single-token comparison that yields a unique, deterministic attachment point.

\paragraph{Component-wise ablation of \algname{}.}
We isolate the contributions of shifted drafting and token matching with a cumulative ablation (Table~\ref{tab:ablation_reuse_matching}). Starting from DFlash, which uses only the native anchor and draft block, we first add shifted blocks drafted from the retained rejected-suffix states, attaching each block to the baseline at its original offset \(t=d\). We then replace this fixed attachment with token matching, which attaches each block at the earliest positive-overlap position. All configurations share the same offsets \(\mathcal D\), drafter evaluations, and verification rule; only the proposal assembly differs. Shifted drafting alone raises \(\tau\) on every benchmark, from 5.38 to 5.87 on average, showing that rejected verifier states provide useful conditioning. Token matching further raises it to 6.22, accounting for roughly 40\% of the total gain, as many shifted continuations become usable only at a position other than their original offset (Table~\ref{tab:token_alignment}).

\section{Conclusion}
We introduced \algname{}, a training-free method that reuses verifier hidden states discarded after rejection to reanchor block diffusion drafting at later positions. By drafting multiple fixed-length blocks in parallel and attaching their continuations through token matching, \algname{} refines predictions within the original block and extends draft coverage beyond its endpoint, while preserving lossless decoding. It consistently improves acceptance length and decoding speed across target models and benchmarks, and applies directly to DFlash-based drafter variants, including Domino, DSpark, and Draft-OPD. These results show that otherwise discarded verifier computations can be effectively reused to improve speculative decoding.

\section*{Limitations}

Like speculative decoding in general, \algname{} is limited by how well the drafter aligns with the target model. 
Shifted blocks are drafted from verifier states conditioned on rejected tokens, so their benefit depends on how much the post-rejection drafts still agree with the target's continuation. The gains therefore vary across tasks and decoding settings. 
At \(T=0\), the average speedup gain is 9.9\% on MT-Bench and Alpaca, compared with 14.5\% on math and code tasks.
Under stochastic sampling (\(T=1\)), independently sampled drafts agree less often with the target, and \algname{} is slower than the baseline on four Llama-3.1-8B tasks and on LiveCodeBench with Qwen3-4B, where the additional verification cost outweighs the extra accepted tokens.
In addition, \algname{} has no retained states to reuse after a round in which the entire block is accepted, so it cannot extend the draft in consecutive full-acceptance rounds. Finally, all latency measurements use batch size one, where verifying additional trie tokens is nearly free; at larger batch sizes, where verification becomes compute-bound, the extra candidates may reduce the net speedup.
\FloatBarrier
\bibliography{references.bib}
\clearpage
\appendix
\section{Variants of DFlash}
This section provides brief descriptions and pseudocode for the baseline methods used in our experiments.
We summarize the core inference procedures of DSpark, Domino and the post-training update of Draft-OPD.
\label{variant_dflash}
\paragraph{DSpark.}
DSpark~\citep{cheng2026dspark} combines parallel drafting with a low-rank Markov correction conditioned on the preceding sampled token. 
A calibrated confidence head estimates conditional acceptance probabilities, whose cumulative products give prefix-survival estimates. 
A hardware-aware scheduler selects the verification length by balancing expected token progress against profiled verification cost.
Let $K\in\{0,\ldots,B-1\}$ be the selected verification length.
For the default Markov variant, $\mathbf h_j^{q,[i]}$ and $\mathbf o_j^{[i]}$ denote the backbone feature and base logits for predicted token position $j$, following DSpark's shifted input--output alignment.
$E_M$ and $W_M$ parameterize the Markov correction, and $\mathrm{Conf}_{j-i}$ includes position-wise calibration.
We summarize scheduling at the block level: $\operatorname{SchedulePrefix}$ uses the hardware profile and current verification workload, extends verification prefixes while estimated throughput improves, and stops at the first non-improving extension.

\begin{algorithm}[htbp]
  \caption{DSpark: drafting and confidence-scheduled verification}
  \label{alg:baseline-dspark}
  \small
  \begin{algorithmic}[1]
    \State \textbf{Input:} $p$; $q$; $\mathbf C_{<i}$; $x_i$;
    $B-1$; hardware profile and current scheduler state.
    \State Compute $(\mathbf h_j^{q,[i]},\mathbf
    o_j^{[i]})_{j=i+1}^{i+B-1}$ in parallel.
    \State Set $\hat x_i^{[i]}\gets x_i$.
    \For{$j=i+1,\ldots,i+B-1$}
    \State $\mathbf
    q_j^{[i]}\gets\operatorname{softmax}\!\left(\mathbf o_j^{[i]}+W_M
    E_M(\hat x_{j-1}^{[i]})\right)$.
    \State $c_j^{[i]}\gets\operatorname{Conf}_{j-i}\!\left(\mathbf
    h_j^{q,[i]},E_M(\hat x_{j-1}^{[i]})\right)$.
    \State Sample $\hat x_j^{[i]}\sim\operatorname{Cat}(\mathbf q_j^{[i]})$.
    \EndFor
    \State For $j=i+1,\ldots,i+B-1$, compute
    \Statex
    \hspace{\algorithmicindent}$a_j^{[i]}\gets\prod_{t=i+1}^{j}c_t^{[i]}$.
    \State $
    \begin{aligned}[t]
      K\gets{}&\operatorname{SchedulePrefix}\!\bigl((a_j^{[i]})_{j=i+1}^{i+B-1};\\
      &\text{hardware profile, scheduler state}\bigr).
    \end{aligned}$
    \State Verify only $\hat{\mathbf x}_{i+1:i+K}^{[i]}$ with $p$.
    \State \Return the accepted prefix and a target-issued token.
  \end{algorithmic}
\end{algorithm}

\paragraph{Domino.}
Domino~\citep{huang2026domino} adds a lightweight causal correction head to a parallel DFlash backbone. A GRU summarizes the anchor and preceding draft tokens, and a low-rank head combines this state with the current backbone feature to correct the base logits. The backbone and full LM head are evaluated once per block; only correction and sampling are sequential.
Let $\mathbf h_j^{q,[i]}$ be the prediction-aligned drafter feature and $\mathbf o_j^{[i]}=W_{\mathrm{LM}}\mathbf h_j^{q,[i]}$ the base logits, where $W_{\mathrm{LM}}$ is the frozen target LM head.
The sequence $\mathbf E_{i:j-1}^{[i]}$ contains embeddings of $x_i$ and the previously sampled tokens $\hat{\mathbf x}_{i+1:j-1}^{[i]}$; $\operatorname{GRU}(\mathbf E_{i:j-1}^{[i]})$ denotes the incrementally maintained recurrent state. The matrices $W_1,W_2$ form the low-rank correction head.

\begin{algorithm}[htbp]
  \caption{Domino: one inference round}
  \label{alg:baseline-domino}
  \small
  \begin{algorithmic}[1]
    \State \textbf{Input:} Target $p$; drafter $q$; context
    $\mathbf C_{<i}$; anchor $a_i$; draft length $B-1$.
    \State Compute $(\mathbf h_j^{q,[i]},\mathbf
    o_j^{[i]})_{j=i+1}^{i+B-1}$ in one parallel backbone and LM-head pass.
    \For{$j=i+1,\ldots,i+B-1$}
    \State $\mathbf s_{j-1}^{[i]}\gets\operatorname{GRU}(\mathbf
    E_{i:j-1}^{[i]})$.
    \State $\Delta\mathbf o_j^{[i]}\gets
    W_2\operatorname{SiLU}\!\left(W_1[\mathbf h_j^{q,[i]};\mathbf
    s_{j-1}^{[i]}]\right)$.
    \State $\mathbf q_j^{[i]}\gets\operatorname{softmax}(\mathbf
    o_j^{[i]}+\Delta\mathbf o_j^{[i]})$.
    \State Sample $\hat x_j^{[i]}\sim\operatorname{Cat}(\mathbf q_j^{[i]})$.
    \EndFor
    \State Verify $\hat{\mathbf x}_{i+1:i+B-1}^{[i]}$ with $p$
    using $(\mathbf q_j^{[i]})$.
    \State \Return the accepted prefix and a target-issued token.
  \end{algorithmic}
\end{algorithm}

\paragraph{Draft-OPD.}
Draft-OPD~\citep{lei2026draftopdonpolicydistillationspeculative} post-trains the drafter through \emph{on-policy distillation}: the target supervises proposals generated by the current draft policy, rather than only fixed target-generated sequences. 
Target-assisted speculative decoding keeps the overall rollout stable while recording draft-block starting positions, proposals, and verification outcomes.
Training returns to these saved block-start contexts to score the drafter's proposals, including discarded suffixes.
Let $\mathcal{I}_{\mathrm{acc}}$ and $\mathcal{I}_{\mathrm{rej}}$ collect accepted and discarded positions across the replayed blocks. $\mathrm{ForwardKL}$ computes the mean of $D_{\mathrm{KL}}(p_j^{[i]}\|q_j^{[i]})$ over accepted positions.
$\mathrm{ReverseKL}$ computes the weighted mean of $D_{\mathrm{KL}}(q_j^{[i]}\|p_j^{[i]})$ over discarded positions, using block-position weights $w_{i,j}=\gamma^{j-i-1}$, where $0<\gamma\leq 1$, and normalizing by the sum of the discarded-position weights. Empty groups contribute zero. Inference remains unchanged.

\begin{algorithm}[htbp]
  \caption{Draft-OPD: one on-policy training update of a DFlash drafter}
  \label{alg:baseline-draftopd}
  \small
  \begin{algorithmic}[1]
    \State \textbf{Input:} Prompts $\mathcal X$; frozen target $p$;
    current drafter $q$; $B-1$; learning rate $\eta$; decay factor $\gamma$.
    \State Run target-assisted speculative rollouts with $q$;
    record block starts, contexts, proposals, and verification outcomes.
    \For{each recorded draft block starting at $i$}
    \State Recompute $(\mathbf q_j^{[i]})_{j=i+1}^{i+B-1}$ in
    parallel from the cached context and original anchor $(\mathbf C_{<i},x_i)$.
    \State Score the recorded proposal with $p$ at each position $j$:
    \Statex \hspace{\algorithmicindent}$p_j^{[i]}\gets
    p\!\left(\,\cdot\mid x_{\leq i},\hat{\mathbf x}_{i+1:j-1}^{[i]}\right)$.
    \EndFor
    \State Collect $\mathcal I_{\mathrm{acc}}$ and $\mathcal
    I_{\mathrm{rej}}$; the latter includes the first rejection and
    all subsequent positions in its block.
    \State $\mathcal
    L_{\mathrm{acc}}\gets\operatorname{ForwardKL}(\mathcal I_{\mathrm{acc}})$.
    \State $\mathcal{L}_{\mathrm{rej}}
    \leftarrow \mathrm{ReverseKL}(\mathcal{I}_{\mathrm{rej}};\gamma)$.
    \State $\theta\gets\theta-\eta\nabla\big[(\mathcal
    L_{\mathrm{acc}}+\mathcal L_{\mathrm{rej}})/2\big]$.
    \Statex Keep $p$ and the sampled proposals fixed.
  \end{algorithmic}
\end{algorithm}

\section{Experiment Details}
\label{exp_detail}
All implementations for DFlash and its variants are based on the publicly available implementations.
We implement inference in PyTorch with Hugging Face Transformers and run both the target and draft models in bfloat16 using PyTorch SDPA. Due to the nature of \algname{} that constructs a trie for every decoding round, shifted proposals are not processed in Python but passed in a native C++17 backend, compiled with \texttt{g++ -O3} as a shared library. The backend aligns each proposal at its longest matching prefix, merges the resulting paths into a token trie, and prepares node, position, and ancestor-attention data; Python transfers these inputs to the GPU for joint verification in one target-model pass. End-to-end latency includes per-round trie construction and preparation of verifier inputs, but excludes model loading, tokenization, and compilation of the C++ backend.
\section{Additional experiments}
\label{more_exp}

\subsection{Winning rate of earliest attachment selection}
\label{app:earliest_attachment}
We evaluate the earliest positive-overlap rule with Qwen3-4B as the target and DFlash as the drafter. We use greedy decoding (\(T=0\)) with a 128-token generation cap and evaluate 100 randomly shuffled examples per task (seed 0) from GSM8K, HumanEval, and MBPP. For each verifier round and shifted offset, we jointly verify paths from all positive-overlap sites together with the baseline path. A group is eligible if at least one attached path ties the maximum accepted draft length among all verified paths. The earliest site counts as a win when its path ties that maximum; ties are included. The table reports the number of eligible groups, the number won by the earliest site, and the corresponding fraction.

\begin{table}[t]
  \centering
  \small
  \begin{tabular*}{\columnwidth}{@{\extracolsep{\fill}}lll}
    \toprule
    \# Tested draft groups & \# Earliest wins & Win rate \\
    \midrule
    25,304 & 23,875 & 94.35\% \\
    \bottomrule
  \end{tabular*}
    \caption{\textbf{Winning rate of earliest attachment selection.} Each tested
    draft group is one verifier round and shifted offset with at least
    one attachment path tying the maximum accepted draft length among
    all verified paths, including the baseline. ``Earliest wins'' counts
    groups where the earliest positive-overlap site also ties this
    maximum; ``Win rate'' is this count divided by the number of tested
    groups.}
  \label{tab:earliest_attachment_winning_rate}
\end{table}

\section{Related Work}
\label{related_work}
\textbf{Speculative Decoding.}
Autoregressive LLMs achieve strong performance but are often noted for their slow decoding speed. While generation must proceed sequentially, verification of multiple tokens can be done in parallel through batched computation~\citep{leviathan2023fast,
chen2023accelerating}. Speculative decoding exploits this property, using a smaller model (the drafter) to propose tokens and the large model to verify them. Spending more computation on the drafter raises acceptance but slows drafting, so the drafter must align with the verifier under a tight compute budget. SpecInfer~\citep{miao2024specinfer} introduced the tree attention mechanism, which is used broadly by subsequent methods as an almost plug-and-play component. Following \citet{stern2018blockwise}, Medusa~\citep{cai2024medusa} brings the drafter into the verifier model using multiple decoding heads, since the computation over the prefix has already been done during verification. EAGLE~\citep{li2024eagle} instead uses a feature-level autoregressive drafter so that the draft carries causality, and its successors add dynamic draft trees~\citep{li2024eagle2} and multi-layer feature fusion~\citep{li2026eagle}.

\textbf{Diffusion Language Model.}
Masked diffusion language models generate tokens in any order and in parallel, committing multiple positions per forward pass. Capability has scaled from D3PM~\citep{austin2021structured} through continuous-time masked formulations~\citep{lou2023discrete,sahoo2024simple, shi2024simplified} to billion-parameter models such as LLaDA \citep{nie2026large} and Dream \citep{ye2025dream7bdiffusionlarge}. Despite this progress, dLLMs have not replaced autoregressive models. Committing more tokens per step degrades quality, since tokens decoded in the same step do not condition on one another~\citep{liu2025discrete}, and the arbitrary decoding order with bidirectional attention prevents KV cache reuse~\citep{wu2026fast}.
Block diffusion~\citep{arriola2025block} interpolates between the two, using bidirectional attention within a block and causal attention across blocks. This recovers cache reuse at block granularity and yields a block-parallel proposal structure that closely mirrors speculative decoding~\citep{leviathan2023fast,
chen2023accelerating}.

\textbf{Speculative Decoding with Diffusion Draft Model.}
Several methods supply drafts through parallel masked prediction, either by sampling candidate paths from a diffusion model and scoring them before committing, or by adapting a small independent model to predict multiple positions at once~\citep{li2026diffuspec,
an2026pard}. DFlash~\citep{chen2026dflash} instead conditions the drafter on the target's hidden states, following EAGLE~\citep{li2024eagle}. Subsequent methods weaken the conditional independence inside a block by imposing causal structure among drafted positions~\citep{huang2026domino, cheng2026dspark}. D²SD~\citep{zhang2026d} truncates the draft and re-runs the drafter from the accepted prefix, so that later positions are proposed under stronger conditioning. Without retraining the drafter, another line reallocates the verification budget, expanding the per-position marginals into a candidate tree verified in one pass. DDTree~\citep{ringel2026accelerating} maximizes expected acceptance length under a fixed node count, and later work makes the budget adaptive by folding verification cost into the objective~\citep{zhang2026cost,
oh2026bastion}.
\end{document}

%% file: math_commands.tex
\usepackage{amsmath,amsfonts,bm}

\def\eqref#1{equation~\ref{#1}}

\def\1{\bm{1}}

\DeclareMathAlphabet{\mathsfit}{\encodingdefault}{\sfdefault}{m}{sl}
\SetMathAlphabet{\mathsfit}{bold}{\encodingdefault}{\sfdefault}{bx}{n}



%% file: references.bib
@inproceedings{leviathan2023fast,
  title={Fast inference from transformers via speculative decoding},
  author={Leviathan, Yaniv and Kalman, Matan and Matias, Yossi},
  booktitle={International conference on machine learning},
  pages={19274--19286},
  year={2023},
  organization={PMLR}
}

@article{chen2023accelerating,
  title={Accelerating large language model decoding with speculative sampling},
  author={Chen, Charlie and Borgeaud, Sebastian and Irving, Geoffrey and Lespiau, Jean-Baptiste and Sifre, Laurent and Jumper, John},
  journal={arXiv preprint arXiv:2302.01318},
  year={2023}
}

@inproceedings{miao2024specinfer,
  title={Specinfer: Accelerating large language model serving with tree-based speculative inference and verification},
  author={Miao, Xupeng and Oliaro, Gabriele and Zhang, Zhihao and Cheng, Xinhao and Wang, Zeyu and Zhang, Zhengxin and Wong, Rae Ying Yee and Zhu, Alan and Yang, Lijie and Shi, Xiaoxiang and others},
  booktitle={Proceedings of the 29th ACM International Conference on Architectural Support for Programming Languages and Operating Systems, Volume 3},
  pages={932--949},
  year={2024}
}

@article{stern2018blockwise,
  title={Blockwise parallel decoding for deep autoregressive models},
  author={Stern, Mitchell and Shazeer, Noam and Uszkoreit, Jakob},
  journal={Advances in Neural Information Processing Systems},
  volume={31},
  year={2018}
}

@article{cai2024medusa,
  title={Medusa: Simple llm inference acceleration framework with multiple decoding heads},
  author={Cai, Tianle and Li, Yuhong and Geng, Zhengyang and Peng, Hongwu and Lee, Jason D and Chen, Deming and Dao, Tri},
  journal={arXiv preprint arXiv:2401.10774},
  year={2024}
}

@article{li2024eagle,
  title={Eagle: Speculative sampling requires rethinking feature uncertainty},
  author={Li, Yuhui and Wei, Fangyun and Zhang, Chao and Zhang, Hongyang},
  journal={arXiv preprint arXiv:2401.15077},
  year={2024}
}

@inproceedings{li2024eagle2,
  title={Eagle-2: Faster inference of language models with dynamic draft trees},
  author={Li, Yuhui and Wei, Fangyun and Zhang, Chao and Zhang, Hongyang},
  booktitle={Proceedings of the 2024 conference on empirical methods in natural language processing},
  pages={7421--7432},
  year={2024}
}

@article{li2026eagle,
  title={Eagle-3: Scaling up inference acceleration of large language models via training-time test},
  author={Li, Yuhui and Wei, Fangyun and Zhang, Chao and Zhang, Hongyang},
  journal={Advances in Neural Information Processing Systems},
  volume={38},
  pages={136737--136756},
  year={2026}
}

@article{austin2021structured,
  title={Structured denoising diffusion models in discrete state-spaces},
  author={Austin, Jacob and Johnson, Daniel D and Ho, Jonathan and Tarlow, Daniel and Van Den Berg, Rianne},
  journal={Advances in neural information processing systems},
  volume={34},
  pages={17981--17993},
  year={2021}
}

@article{lou2023discrete,
  title={Discrete diffusion modeling by estimating the ratios of the data distribution},
  author={Lou, Aaron and Meng, Chenlin and Ermon, Stefano},
  journal={arXiv preprint arXiv:2310.16834},
  year={2023}
}

@article{sahoo2024simple,
  title={Simple and effective masked diffusion language models},
  author={Sahoo, Subham S and Arriola, Marianne and Schiff, Yair and Gokaslan, Aaron and Marroquin, Edgar and Chiu, Justin T and Rush, Alexander and Kuleshov, Volodymyr},
  journal={Advances in Neural Information Processing Systems},
  volume={37},
  pages={130136--130184},
  year={2024}
}

@article{shi2024simplified,
  title={Simplified and generalized masked diffusion for discrete data},
  author={Shi, Jiaxin and Han, Kehang and Wang, Zhe and Doucet, Arnaud and Titsias, Michalis},
  journal={Advances in neural information processing systems},
  volume={37},
  pages={103131--103167},
  year={2024}
}

@article{nie2026large,
  title={Large language diffusion models},
  author={Nie, Shen and Zhu, Fengqi and You, Zebin and Zhang, Xiaolu and Ou, Jingyang and Hu, Jun and Zhou, Jun and Lin, Yankai and Wen, Ji-Rong and Li, Chongxuan},
  journal={Advances in Neural Information Processing Systems},
  volume={38},
  pages={50608--50646},
  year={2026}
}

@misc{ye2025dream7bdiffusionlarge,
      title={Dream 7B: Diffusion Large Language Models}, 
      author={Jiacheng Ye and Zhihui Xie and Lin Zheng and Jiahui Gao and Zirui Wu and Xin Jiang and Zhenguo Li and Lingpeng Kong},
      year={2025},
      eprint={2508.15487},
      archivePrefix={arXiv},
      primaryClass={cs.CL},
      url={https://arxiv.org/abs/2508.15487}, 
}

@inproceedings{liu2025discrete,
  title={Discrete copula diffusion},
  author={Liu, Anji and Broadrick, Oliver and Niepert, Mathias and Van den Broeck, Guy},
  booktitle={International Conference on Learning Representations},
  volume={2025},
  pages={88953--88979},
  year={2025}
}

@inproceedings{wu2026fast,
  title={Fast-dllm: Training-free acceleration of diffusion llm by enabling kv cache and parallel decoding},
  author={Wu, Chengyue and Zhang, Hao and Xue, Shuchen and Liu, Zhijian and Diao, Shizhe and Zhu, Ligeng and Luo, Ping and Han, Song and Xie, Enze},
  booktitle={International Conference on Learning Representations},
  volume={2026},
  pages={57027--57051},
  year={2026}
}

@inproceedings{arriola2025block,
  title={Block diffusion: Interpolating between autoregressive and diffusion language models},
  author={Arriola, Marianne and Gokaslan, Aaron and Chiu, Justin and Yang, Zhihan and Qi, Zhixuan and Han, Jiaqi and Sahoo, Subham and Kuleshov, Volodymyr},
  booktitle={International Conference on Learning Representations},
  volume={2025},
  pages={50726--50753},
  year={2025}
}

@inproceedings{li2026diffuspec,
  title={Diffuspec: Unlocking diffusion language models for speculative decoding},
  author={Li, Guanghao and Fu, Zhihui and Fang, Min and Zhao, Qibin and Tang, Ming and Yuan, Chun and Wang, Jun},
  booktitle={Findings of the Association for Computational Linguistics: ACL 2026},
  pages={20896--20910},
  year={2026}
}

@inproceedings{an2026pard,
  title={PARD: Accelerating LLM Inference with Low-Cost PARallel Draft Model Adaptation},
  author={An, Zihao and Bai, Huajun and Liu, Ziqiong and Li, Dong and Barsoum, Emad},
  booktitle={International Conference on Learning Representations},
  volume={2026},
  pages={144562--144579},
  year={2026}
}

@article{chen2026dflash,
  title={Dflash: Block diffusion for flash speculative decoding},
  author={Chen, Jian and Liang, Yesheng and Liu, Zhijian},
  journal={arXiv preprint arXiv:2602.06036},
  year={2026}
}

@article{huang2026domino,
  title={Domino: Decoupling causal modeling from autoregressive drafting in speculative decoding},
  author={Huang, Jianuo and Zhang, Yaojie and Zhang, Qituan and Lin, Hao and Xu, Hanlin and Zhang, Linfeng},
  journal={arXiv preprint arXiv:2605.29707},
  year={2026}
}

@article{cheng2026dspark,
  title={DSpark: Confidence-Scheduled Speculative Decoding with Semi-Autoregressive Generation},
  author={Cheng, Xin and Yu, Xingkai and Shao, Chenze and Li, Jiashi and Xiong, Yunfan and Qian, Yi and Zhu, Jiaqi and Ma, Shirong and Zhang, Xiaokang and Ye, Jiasheng and others},
  journal={arXiv preprint arXiv:2607.05147},
  year={2026}
}

@article{zhang2026d,
  title={D\^{} 2SD: Accelerating Speculative Decoding with Dual Diffusion Draft Models},
  author={Zhang, Liyuan and Zhang, Jiarui and Yao, Jinwei and Yan, Ran and Yang, Yuchen and Zhang, Jiahao and Yang, Tongkai and Wu, Yi and Yuan, Binhang},
  journal={arXiv preprint arXiv:2606.04446},
  year={2026}
}

@article{ringel2026accelerating,
  title={Accelerating speculative decoding with block diffusion draft trees},
  author={Ringel, Liran and Romano, Yaniv},
  journal={arXiv preprint arXiv:2604.12989},
  year={2026}
}

@article{oh2026bastion,
  title={Bastion: Budget-Aware Speculative Decoding with Tree-structured Block Diffusion Drafting},
  author={Oh, Soowon and Cao, Nam and Kim, Yujin and Jung, Hojung and Ahmad, Huzama and Bae, Sangmin and Yun, Se-Young},
  journal={arXiv preprint arXiv:2605.29727},
  year={2026}
}

@article{zhang2026cost,
  title={Cost-Aware Diffusion Draft Trees for Speculative Decoding},
  author={Zhang, Shuai and Qiu, Huachuan and He, Hongliang and Dai, Yong},
  journal={arXiv preprint arXiv:2606.01813},
  year={2026}
}

@article{cheng2024recurrent,
  title={Recurrent drafter for fast speculative decoding in large language models},
  author={Cheng, Yunfei and Zhang, Aonan and Zhang, Xuanyu and Wang, Chong and Wang, Yi},
  journal={arXiv preprint arXiv:2403.09919},
  year={2024}
}

@inproceedings{zhang2025draft,
  title={Draft model knows when to stop: Self-verification speculative decoding for long-form generation},
  author={Zhang, Ziyin and Xu, Jiahao and Liang, Tian and Chen, Xingyu and He, Zhiwei and Wang, Rui and Tu, Zhaopeng},
  booktitle={Proceedings of the 2025 Conference on Empirical Methods in Natural Language Processing},
  pages={16696--16708},
  year={2025}
}

@article{huang2024specdec,
  title={Specdec++: Boosting speculative decoding via adaptive candidate lengths},
  author={Huang, Kaixuan and Guo, Xudong and Wang, Mengdi},
  journal={arXiv preprint arXiv:2405.19715},
  year={2024}
}

@article{cobbe2021training,
  title={Training verifiers to solve math word problems},
  author={Cobbe, Karl and Kosaraju, Vineet and Bavarian, Mohammad and Chen, Mark and Jun, Heewoo and Kaiser, Lukasz and Plappert, Matthias and Tworek, Jerry and Hilton, Jacob and Nakano, Reiichiro and others},
  journal={arXiv preprint arXiv:2110.14168},
  year={2021}
}

@inproceedings{lightman2024let,
  title={Let's verify step by step},
  author={Lightman, Hunter and Kosaraju, Vineet and Burda, Yuri and Edwards, Harrison and Baker, Bowen and Lee, Teddy and Leike, Jan and Schulman, John and Sutskever, Ilya and Cobbe, Karl},
  booktitle={International Conference on Learning Representations},
  volume={2024},
  pages={39578--39601},
  year={2024}
}

@article{chen2021evaluating,
  title={Evaluating large language models trained on code},
  author={Chen, Mark and Tworek, Jerry and Jun, Heewoo and Yuan, Qiming and Pinto, Henrique Ponde De Oliveira and Kaplan, Jared and Edwards, Harri and Burda, Yuri and Joseph, Nicholas and Brockman, Greg and others},
  journal={arXiv preprint arXiv:2107.03374},
  year={2021}
}

@article{austin2021program,
  title={Program synthesis with large language models},
  author={Austin, Jacob and Odena, Augustus and Nye, Maxwell and Bosma, Maarten and Michalewski, Henryk and Dohan, David and Jiang, Ellen and Cai, Carrie and Terry, Michael and Le, Quoc and others},
  journal={arXiv preprint arXiv:2108.07732},
  year={2021}
}

@inproceedings{jain2025livecodebench,
  title={Livecodebench: Holistic and contamination free evaluation of large language models for code},
  author={Jain, Naman and Gu, Alex and Li, Wen-Ding and Yan, Fanjia and Zhang, Tianjun and Wang, Sida and Solar-Lezama, Armando and Sen, Koushik and Stoica, Ion},
  booktitle={International Conference on Learning Representations},
  volume={2025},
  pages={58791--58831},
  year={2025}
}

@article{zheng2023judging,
  title={Judging llm-as-a-judge with mt-bench and chatbot arena},
  author={Zheng, Lianmin and Chiang, Wei-Lin and Sheng, Ying and Zhuang, Siyuan and Wu, Zhanghao and Zhuang, Yonghao and Lin, Zi and Li, Zhuohan and Li, Dacheng and Xing, Eric and others},
  journal={Advances in neural information processing systems},
  volume={36},
  pages={46595--46623},
  year={2023}
}

@article{taori2023alpaca,
  title={Alpaca: A strong, replicable instruction-following model},
  author={Taori, Rohan and Gulrajani, Ishaan and Zhang, Tianyi and Dubois, Yann and Li, Xuechen and Guestrin, Carlos and Liang, Percy and Hashimoto, Tatsunori B},
  journal={Stanford Center for Research on Foundation Models. https://crfm. stanford. edu/2023/03/13/alpaca. html},
  volume={3},
  number={6},
  pages={7},
  year={2023}
}

@misc{lei2026draftopdonpolicydistillationspeculative,
      title={Draft-OPD: On-Policy Distillation for Speculative Draft Models}, 
      author={Haodi Lei and Yafu Li and Haoran Zhang and Shunkai Zhang and Qianjia Cheng and Xiaoye Qu and Ganqu Cui and Bowen Zhou and Ning Ding and Yun Luo and Yu Cheng},
      year={2026},
      eprint={2605.29343},
      archivePrefix={arXiv},
      primaryClass={cs.CL},
      url={https://arxiv.org/abs/2605.29343}, 
}

@article{yang2025qwen3,
  title={Qwen3 technical report},
  author={Yang, An and Li, Anfeng and Yang, Baosong and Zhang, Beichen and Hui, Binyuan and Zheng, Bo and Yu, Bowen and Gao, Chang and Huang, Chengen and Lv, Chenxu and others},
  journal={arXiv preprint arXiv:2505.09388},
  year={2025}
}

@article{grattafiori2024llama,
  title={The llama 3 herd of models},
  author={Grattafiori, Aaron and Dubey, Abhimanyu and Jauhri, Abhinav and Pandey, Abhinav and Kadian, Abhishek and Al-Dahle, Ahmad and Letman, Aiesha and Mathur, Akhil and Schelten, Alan and Vaughan, Alex and others},
  journal={arXiv preprint arXiv:2407.21783},
  year={2024}
}

@misc{gemmateam2026gemma4technicalreport,
      title={Gemma 4 Technical Report}, 
      author={Gemma Team and Sherif El Abd and Vaibhav Aggarwal and Robin Algayres and Alek Andreev and Olivier Bachem and Ian Ballantyne and Cormac Brick and Victor Cărbune and Michelle Casbon and Mayank Chaturvedi and Aditya Chawla and Victor Cotruta and Alice Coucke and Phil Culliton and Robert Dadashi and Lucas Dixon and Mohamed Elhawaty and Utku Evci and Clément Farabet and Johan Ferret and Filippo Galgani and Sertan Girgin and Jean-Bastien Grill and Maarten Grootendorst and Jiaxian Guo and Cassidy Hardin and Yanzhang He and Steven M. Hernandez and Omri Homburger and Léonard Hussenot and Juyeong Ji and Armand Joulin and Aishwarya Kamath and Parnian Kassraie and Olivier Lacombe and Preethi Lahoti and Gaël Liu and Gus Martins and Luciano Martins and Tatiana Matejovicova and Ramona Merhej and Nikola Momchev and Sneha Mondal and Ryan Mullins and Sindhu Raghuram Panyam and Shreya Pathak and Sarah Perrin and André Susano Pinto and Etienne Pot and Angéline Pouget and Alexandre Ramé and Sabela Ramos and Douglas Reid and David Rim and Morgane Rivière and Karsten Roth and Louis Rouillard and Omar Sanseviero and Pier Giuseppe Sessa and Shane Settle and Danila Sinopalnikov and Sara Smoot and Piotr Stanczyk and Andreas Steiner and Lawrence Stewart and Ilya Tolstikhin and Michael Tschannen and Anton Tsitsulin and Nino Vieillard and Renjie Wu and Pingmei Xu and Haichuan Yang and Edouard Yvinec and Biao Zhang and Li Zhang and Joe Zou and Nicolas Aagnes and Abdelrahman Abdelhamed and Jakub Adamek and Shivani Agrawal and Shubham Agrawal and Ibrahim Alabdulmohsin and Jean Baptiste Alayrac and Uri Alon and Chandramouli Amarnath and Ankesh Anand and Chrysovalantis Anastasiou and Setareh Ariafar and François-Xavier Aubet and Kyriakos Axiotis and Federico Barbero and Joelle Barral and Alexei Bendebury and Urs Bergmann and Stanley Bileschi and Kat Black and Mathieu Blondel and Sebastian Borgeaud and Arthur Bražinskas and Ryan Burnell and Robert Busa-Fekete and Mu Cai and Daniele Calandriello and Glenn Cameron and Charlotte Caucheteux and Rahma Chaabouni and Garima Chadha and Jetha Chan and Blake Jianhang Chen and Jesse Chen and Lin Chen and Xu Chen and Derek Cheng and Tzu-hsiang Chien and Nikolai Chinaev and Yi Chou and Zhaohui Chu and Benjamin Coleman and Pooja Consul and Sam Conway-Rahman and Scott Crowell and Dylan Cutler and Vivek Dani and Samira Daruki and Anil Das and Daniel Deutsch and Nishanth Dikkala and Li Ding and Qiuhan Ding and Shenil Dodhia and Konstantin Donhauser and Tulsee Doshi and Anca Dragan and Alex Druinsky and Sahil Dua and Zoltan Egyed and Danielle Eisenbud and Daniel Eppens and Cindy Fan and Bahare Fatemi and Yassir Fathullah and Vlad Feinberg and Milen Ferev and Sebastian Flennerhag and Takumi Fujimoto and João Gabriel Oliveira and Isaac Galatzer-Levy and João Gante and Simon Geisler and Soham Ghosal and Antonious M. Girgis and Tamara von Glehn and Alec Go and Alhaad Gokhale and Alex Grills and Yiming Gu and Mayank Gupta and Pramod Gupta and Guru Guruganesh and Raia Hadsell and Hamza Harkous and Jitendra Harlalka and Demis Hassabis and Anja Hauth and Joe Heyward and Arian Hosseini and Chih-Yang Hsia and I-Hung Hsu and Xiaopeng Huang and Yangsibo Huang and Kevin Hui and Adrian Hutter and Te I and Fotis Iliopoulos and Advait Jain and Ganesh Jawahar and Ziwei Ji and Qilin Jin and Melvin Johnson and Kandarp Joshi and Arun Kandoor and Wang-Cheng Kang and Koray Kavukcuoglu and Mehran Kazemi and Kathleen Kenealy and Amr Khalifa and Phoebe Kirk and Ivan Korotkov and Suraj Kothawade and Vitaly Kovalev and Neel Kovelamudi and Adam Kraft and Ravin Kumar and Vivek Kumar and Harish Kuppam and Justin Lannin and Chen-Yu Lee and Seungji Lee and Dmitry Lepikhin and Alon Levkovitch and Dongdong Li and Qiujia Li and Valentin Liévin and Ethan Lin and Ziqian Lin and Casper Liu and Tianlin Liu and Tianqi Liu and Xin Liu and Ivan Lobov and Mayank Lunayach and Min Ma and Gagan Madan and Andrii Maksai and Eric Malmi and Michal Matuszak and Daniel McDuff and Gaurav Menghani and Maciej Mikuła and Daniil Mirylenka and Karolis Misiunas and Vedant Misra and Andreea Mitran and Kareem Mohamed and Maksim Mukha and Eric Noland and James O'Donnell and Brendan O'Donoghue and Kate Olszewska and Bernett Orlando and Wanqiong Pan and Rina Panigrahy and Unnati Parekh and Nicolas Perez-Nieves and Chunjong Park and Eric Paskie and Liqian Peng and Bryce Petrini and Slav Petrov and Jonas Pfeiffer and Bilal Piot and Martyna Plomecka and Siim Poder and Octavio Ponce and Arijit Pramanik and David Racz and Anish Rajan and Michelle Ramanovich and Anand Rao and Marvin Ritter and Vitor Rodrigues and Evan Rosen and Mikołaj Rybiński and Noveen Sachdeva and Michaël E. Sander and Rohit Sathyanarayana and Sagar Savla and Samuel Schmidgall and Tal Schuster and George Scrivener and Benoit Seguin and Andrew Sellergren and Aliaksei Severyn and Izhak Shafran and Dhruv Shah and Bobak Shahriari and Yuan Shangguan and Ashish Shenoy and Pradeep Shenoy and Rakesh Shivanna and Pauline Sho and Lucas Spangher and Wojciech Stokowiec and Tim Strother and Yao Su and Yinghao Sun and Mukund Sundararajan and Andrea Tacchetti and Mor Hazan Taege and Pouya Tafti and Jean Tarbouriech and Chetan Tekur and Shantanu Thakoor and Rahul Thapa and Madeleine Traverse and Lenart Treven and Tao Tu and Chien Te Tung and Çağlar Ünlü and Petar Veličković and Malini Pooni Venkat and Sagar Gubbi Venkatesh and Vidya Venkiteswaran and Francesco Visin and Alex Vitvitskyi and Kiran Vodrahalli and Weiyi Wang and Xin Wang and Tris Warkentin and Jan Wassenberg and John Wieting and Cindy Wu and Lechao Xiao and Hao Xu and Yuhui Xu and Fuzhao Xue and Arun Yadav and Jun Yan and Antoine Yang and Lin Yang and Ming-Hsuan Yang and Ziyu Ying and Jae Hyeon Yoo and Morteza Zadimoghaddam and Sajjad Zafar and Fred Zhang and Jiageng Zhang and Jianyi Zhang and Xiaofan Zhang and Chao Zhao and David Zhou and Chen Zou},
      year={2026},
      eprint={2607.02770},
      archivePrefix={arXiv},
      primaryClass={cs.CL},
      url={https://arxiv.org/abs/2607.02770}, 
}
